\documentclass[letterpaper, 10 pt, conference]{ieeeconf}  

\usepackage{amsmath,amsfonts}
\usepackage{algorithmic}
\usepackage{algorithm}
\usepackage{array}
\usepackage[caption=false,font=normalsize,labelfont=sf,textfont=sf]{subfig}
\usepackage{textcomp}
\usepackage{stfloats}
\usepackage{url}
\usepackage{verbatim}
\usepackage{graphicx}
\usepackage{cite}
\usepackage{xcolor}
\usepackage[left=1.91cm, right = 1.91cm, top=1.91cm, bottom=1.91cm]{geometry}
\let\labelindent\relax
\usepackage{enumitem}
\usepackage{hyperref}
\usepackage{multirow}
\usepackage{float}
\usepackage[font=normalsize,labelfont=bf]{caption}
\def\BibTeX{{\rm B\kern-.05em{\sc i\kern-.025em b}\kern-.08em
    T\kern-.1667em\lower.7ex\hbox{E}\kern-.125emX}}

\IEEEoverridecommandlockouts                              

\title{\LARGE 
\textbf{COMPAct}: \textbf{C}omputational \textbf{O}ptimization and Automated \textbf{M}odular design of 
\textbf{P}lanetary \textbf{Act}uators
}

\author{Aman Singh$^{*,1}$, Deepak Kapa$^{*,2}$, Suryank Joshi$^{1}$, and Shishir Kolathaya$^{1}$%
\thanks{*Authors contributed equally to this work.}%
\thanks{This research is supported by ARTPARK at IISc.}%
\thanks{$^{1}$Department of Cyber-Physical Systems (CPS), Indian Institute of Science (IISc), Bengaluru, India.
{\tt\scriptsize \{saman, suryankjoshi, shishirk\}@iisc.ac.in}}%
\thanks{$^{2}$Department of Physics, Indian Institute of Technology Roorkee, Roorkee, India.
{\tt\scriptsize deepak\_k1@ph.iitr.ac.in}}%
}

\begin{document}

\maketitle
\thispagestyle{empty}
\pagestyle{empty}

\begin{abstract}
The optimal design of robotic actuators is a critical area of research, yet limited attention has been given to optimizing gearbox parameters and automating actuator CAD. This paper introduces \textbf{COMPAct}: \textit{Computational Optimization and Automated Modular Design of Planetary Actuators}, a framework that systematically identifies optimal gearbox parameters for a given motor across four gearbox types, single-stage planetary gearbox (SSPG), compound planetary gearbox (CPG), Wolfrom planetary gearbox (WPG), and double-stage planetary gearbox (DSPG). The framework minimizes mass and actuator width while maximizing efficiency, and further automates actuator CAD generation to enable direct 3D printing without manual redesign. Using this framework, optimal gearbox designs are explored across a wide range of gear ratios, providing insights into the suitability of different gearbox types while automatically generating CAD models for all four gearbox types with varying gear ratios and motors. Two actuator types are fabricated and experimentally evaluated through power efficiency, no-load backlash, and transmission stiffness tests. Experimental results indicate that the SSPG actuator achieves a mechanical efficiency of $60$--$80\%$, a no-load backlash of $0.59^\circ$, and a transmission stiffness of $242.7~\mathrm{Nm/rad}$, while the CPG actuator demonstrates $60\%$ efficiency, $2.6^\circ$ backlash, and a stiffness of $201.6~\mathrm{Nm/rad}$. 
CODE: \href{https://github.com/singhaman1750/COMPAct.git}{Github}
VIDEO: \href{https://youtu.be/etK6anjXag8?si=jFK7HgAPSBy-GnDR}{Supplemental Video}
\end{abstract}


\section{Introduction}

Computational robot design for optimal performance has attracted significant attention, 
particularly through design frameworks that optimize mechanical design for particular application. 
Studies such as \cite{StarlETHCooptimization, Co-designing_versatile_quadruped_robots_for_dynamic_and_energy-efficient_motions, Vitruvio, MetaRLCodesign} optimize link lengths, transmission ratios, and spring stiffness but exclude gearboxes. Similarly, \cite{JointOptIFT} considers link lengths and actuator attachments without optimizing gearbox parameters. 
Other efforts \cite{StochasticProgCodesign1, StochasticProgCodesign2} apply design optimization to manipulators and monopeds, including gear ratios, compliance, and mass, yet without addressing gearbox type, efficiency, or weight. Additional works \cite{A_versatile_co-design___legged_robots, Computational_design_of____size_and_actuators, Simulation_Aided_Co-Design_for_Robust_Robot_Optimization} incorporate motor and gearbox friction but focus on belt drives rather than planetary gears.  
Most approaches thus overlook gearbox design or treat ratios without considering efficiency, mass, or gearbox type, often using simplified mass models. 

\begin{figure}[htbp]
    \centering
    \includegraphics[width=0.8\linewidth]{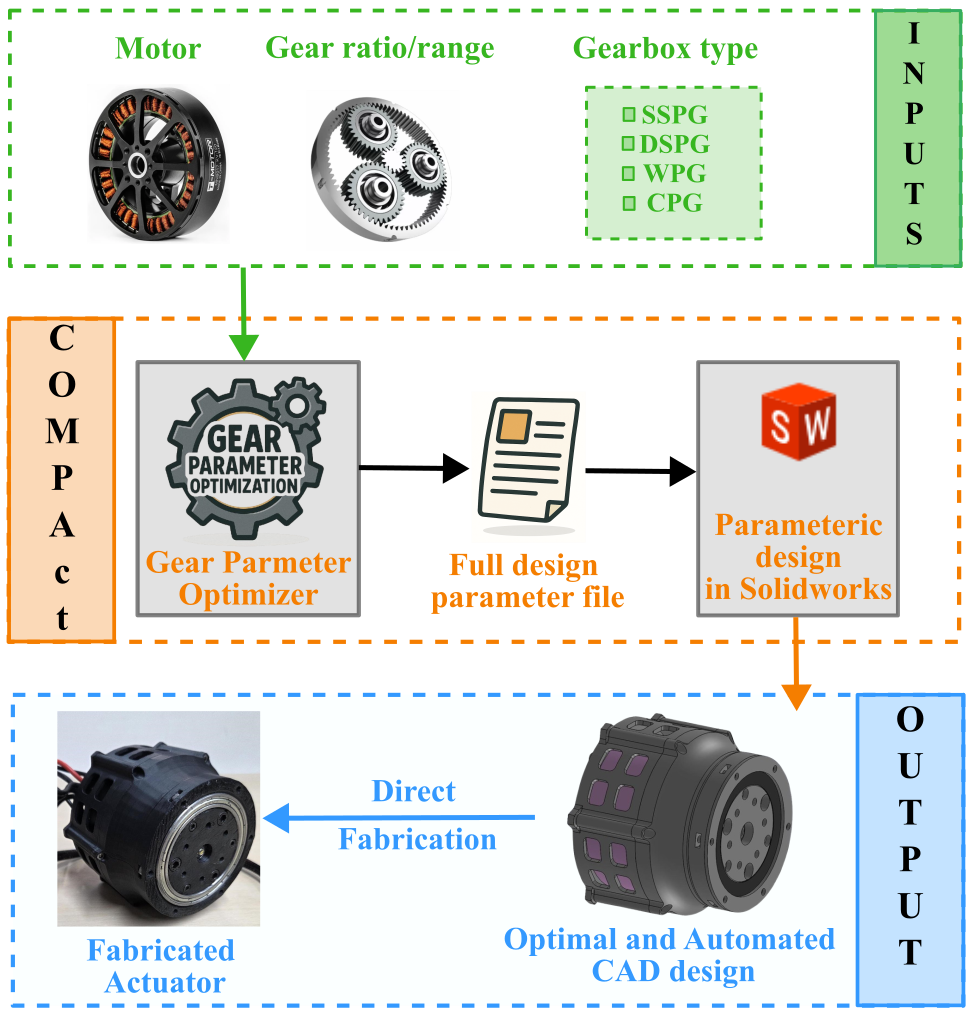}
    \caption{\textbf{COMPAct:} Framework for optimal design and automated CAD generation of planetary robotic actuators.}
    \label{fig:Compact_diag}
\end{figure}

A few works consider gearbox optimization. For example, \cite{motorOpt} optimizes motor–transmission combinations but does not address gearbox type or parameters like teeth and module. Similarly, \cite{PantherLeg, KaistHound} optimize compound and single-stage planetary gearboxes by selecting tooth counts and module while neglecting mass and efficiency. Likewise, \cite{choe2025design3dofhoppingrobot} studies a Wolfrom (3K) gearbox, reducing mass by minimizing tooth counts without detailed modeling or efficiency analysis. The work in \cite{singh2025comparisonexternalinternalsingle} optimizes gearbox parameters but is limited to single-stage planetary actuators.

Beyond optimization, research has explored CAD automation. LLM-based approaches~\cite{wu2021deepcaddeepgenerativenetwork, du2024blenderllmtraininglargelanguage} remain early-stage and cannot yet handle complex systems such as actuators or gearboxes. A more reliable approach is parametric CAD design~\cite{CAMBA201618, ROLLER1991385, CAD_to_URDF, 10.1115/DETC2004-57467, Furst2022}, where geometric dimensions are defined as variables and updates propagate automatically. 
To our knowledge, no prior work has automated the CAD design of planetary gearboxes for robotic actuators.

Alongside these design automation efforts, interest in 3D-printed actuators for robotic joints has grown. While CNC-machined actuators offer higher performance, 3D-printed ones enable rapid prototyping, lower cost, and faster iteration. Prior work has demonstrated their use in planetary actuators~\cite{Embir3DPAct, Exp_3DP}, wearable exoskeletons~\cite{Bezzini3DPforExoskeletons}, humanoid and quadrupedal platforms~\cite{berkeleyhumanoidlite, Solo}, and hobbyist projects~\cite{NTwersky3dpQDDMITminicheetahclone, OpenTorqueActuator}. However, these designs use fixed gear ratios and do not evaluate actuator optimality. A key gap remains in accounting for gearbox mass and efficiency and in automating planetary gearbox design to systematically optimize 3D-printed actuators.
To address these challenges, we present \textbf{COMPAct: Computational Optimization and automated Modular Parametric Design of 3D-Printed Planetary Actuators}, a unified optimization and automation framework, see Fig.\ref{fig:Compact_diag}. The key contributions are:  

\begin{enumerate}
    \item \textbf{Gearbox optimization:} We present an optimization framework for four types of planetary gearboxes. The framework minimizes actuator mass and width while maximizing efficiency using a detailed mass model.
    \item \textbf{CAD automation:} Automated parametric CAD models are generated, enabling direct 3D printing of optimized actuators without requiring manual redesign.  
    \item \textbf{Hardware validation:} Two actuators from the framework, an SSPG (MN8014 motor, 7.2:1) and a CPG (MAD M6C12 motor, 14:1), are fabricated and experimentally tested to validate the proposed approach.      
\end{enumerate}  

The framework supports four gearbox types: Single-Stage (SSPG), Compound (CPG), Wolfrom/3K (WPG), and Double-Stage (DSPG).
The remainder of this paper is organized as follows. Section~\ref{sec:gb_des} introduces the gearbox designs and mass model. Section~\ref{sec:opt} formulates the optimization problem, and Section~\ref{sec:auto} presents the automation framework. Section~\ref{sec:results} reports optimization results, CAD automation, and hardware validation. Finally, Section~\ref{sec:conclusion} concludes the paper and outlines future work.

\section{Gearbox Designs}\label{sec:gb_des}

This section introduces the gearbox architectures analyzed to identify the optimal actuator, along with their gear ratio and efficiency expressions. A unified notation for gear parameters is used for consistency (Section~\ref{subsec:Opt_var}). Only relevant variables are active; e.g., SSPG uses stage-1 parameters only, while CPG and WPG omit certain gears in stage-1 or stage-2. Gear ratios follow standard planetary gear equations~\cite{Theory_of_Machines_and_Mechanism_RS_Khurmi}, and efficiency expressions follow~\cite{3KgearOpt} using basic driving efficiencies from~\cite{BasicDrivingEff}. This efficiency model does not account for gearbox efficiency dependence on gear rotational velocities.

\subsection{Single Stage Planetary Gearbox (SSPG)}

\begin{figure}[htbp]
    \centering
    \includegraphics[width=0.8\linewidth]{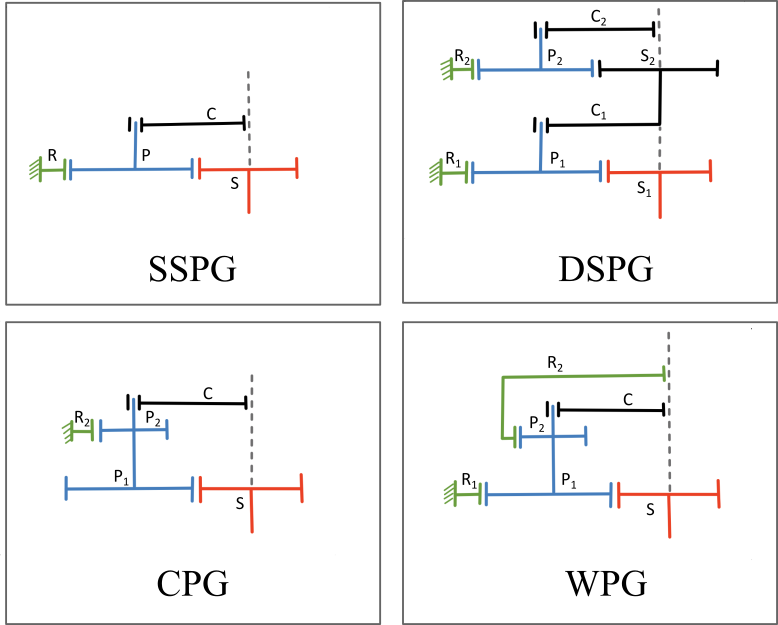}
    \caption{Line diagrams of planetary gearbox configurations. \textbf{SSPG:} Sun ($S$) input, Ring ($R$) fixed, Carrier ($C$) output. \textbf{CPG:} Sun ($S$) input, Ring ($R_2$) fixed, Carrier ($C$) output; planets ($P_1, P_2$) rigidly attached. \textbf{DSPG:} Sun ($S_1$) input, Rings ($R_1, R_2$) fixed, Carrier ($C_2$) output; $S_2$ and $C_1$ rigidly attached. \textbf{WPG (3K):} Sun ($S$) input, Ring ($R_1$) fixed, Ring ($R_2$) output; planets ($P_1, P_2$) rigidly attached.
}
    \label{fig:line_diagram}
\end{figure}

The single-stage planetary gearbox (SSPG) considered in this work has the ring gear fixed, the sun gear driven by the motor, and the carrier providing the output, see Fig. \ref{fig:line_diagram}. This arrangement provides the highest gear reduction among planetary configurations. The gear reduction ratio of the SSPG is given by
\begin{equation}\label{eq:sspg_ratio}
    G_{\text{SSPG}} = \frac{\omega_s}{\omega_c} 
    = \frac{N_{s_1} + N_{r_1}}{N_{s_1}},
\end{equation}
where $G_{\text{SSPG}}$ is the gear reduction ratio, $\omega_c$ is the angular velocity of the carrier (output) shaft, $\omega_s$ is the angular velocity of the sun gear (input) shaft, $N_{s_1}$ is the number of teeth on the sun gear, and $N_{r_1}$ is the number of teeth on the ring gear.  

The efficiency of the SSPG is expressed as
\begin{equation}\label{eq:sspg_eff}
    \eta_{\text{SSPG}} = 
    \frac{N_{s_1} + \eta_{s_1p_1}\,\eta_{p_1r_1}\,N_{r_1}}
         {N_{s_1} + N_{r_1}},
\end{equation}
where $\eta_{\text{SSPG}}$ is the overall efficiency of the gearbox, $\eta_{s_1p_1}$ denotes the efficiency of the sun--planet gear mesh, and $\eta_{p_1r_1}$ denotes the efficiency of the planet--ring gear mesh.  

\subsection{Compound Planetary Gearbox (CPG)}
The compound planetary gearbox (CPG) considered here uses compound planet gears, each comprising two rigidly connected gears of different diameters. In this configuration, the sun gear drives the larger planet, while the smaller planet meshes with a fixed ring gear (Fig.~\ref{fig:line_diagram}). For simplicity, the sun and larger planet module $m_1$, is equal to the smaller planet and ring module $m_2$.

The gear reduction ratio of the CPG is
\begin{equation}\label{eq:cpg_ratio}
    G_{\text{CPG}} = \frac{(N_{s_1} + N_{p_1})(N_{p_2} + N_{p_1})}{N_{s_1} N_{p_2}},
\end{equation}
where $G_{\text{CPG}}$ is the gear reduction ratio, $N_{s_1}$ is the number of teeth on the sun gear, $N_{p_1}$ and $N_{p_2}$ are the numbers of teeth on the larger and smaller planet gears, respectively, and $N_{r_2}$ is the number of teeth on the ring gear.  

The efficiency of the CPG is
\begin{equation}\label{eq:cpg_eff}
    \eta_{\text{CPG}} =
    \frac{N_{s_1} N_{p_2} + 
    \eta_{s_1p_1}\,\eta_{p_2r_2}\,N_{p_1} N_{r_2}}
    {(N_{s_1} + N_{p_1})(N_{p_2} + N_{p_1})},
\end{equation}
where $\eta_{\text{CPG}}$ is the overall efficiency, $\eta_{s_1p_1}$ is the sun--planet mesh efficiency, and $\eta_{p_2r_2}$ is the planet--ring mesh efficiency.  

\subsection{Double-Stage Planetary Gearbox (DSPG)}
The double-stage planetary gearbox (DSPG) employs two successive planetary layers to achieve high gear reductions, see Fig. \ref{fig:line_diagram}. Although less efficient due to the added stages, it is included in this work for its ability to provide exceptionally large gear ratios.  

The gear reduction ratio of the DSPG is
\begin{equation}\label{eq:dspg_ratio}
    G_{\text{DSPG}} 
    = G_{\text{DSPG}}^{1}\,G_{\text{DSPG}}^{2} 
    = \prod_{i=1}^{2}\frac{N_{s_i} + N_{r_i}}{N_{s_i}},
\end{equation}
where $G_{\text{DSPG}}$ is the overall gear reduction ratio, $G_{\text{DSPG}}^{i}$ is the ratio of the $i^{\text{th}}$ stage, $N_{s_i}$ is the number of teeth on the sun gear, and $N_{r_i}$ is the number of teeth on the ring gear in stage $i$.  
The overall efficiency of the DSPG is
\begin{equation}\label{eq:dspg_eff}
    \eta_{\text{DSPG}} 
    = \eta_{\text{DSPG}}^{1}\,\eta_{\text{DSPG}}^{2} 
    = \prod_{i=1}^{2}\frac{N_{s_i} + \eta_{s_ip_i}\,\eta_{p_ir_i}\,N_{r_i}}
    {N_{s_i} + N_{r_i}},
\end{equation}
where $\eta_{\text{DSPG}}$ is the overall efficiency, $\eta_{\text{DSPG}}^{i}$ the efficiency of the $i^{\text{th}}$ stage, $\eta_{s_ip_i}$ the sun--planet mesh efficiency, and $\eta_{p_ir_i}$ the planet--ring mesh efficiency.

\subsection{Wolfrom (3K) Planetary Gearbox (WPG)}\label{subsec:wpg_gb}

The Wolfrom (3K) planetary gearbox (WPG)~\cite{3KgearOpt} is a specialized transmission topology designed to achieve very high gear reductions within a compact volume. Its architecture consists of a sun gear ($S$) driven by the motor, two rigidly connected planet gears ($P_1$ and $P_2$), and two concentric ring gears ($R_1$ and $R_2$). In the configuration considered in this work, the input is applied to the sun gear $S$, the first ring gear $R_1$ is fixed, and the second ring gear $R_2$ provides the output, see Fig. \ref{fig:line_diagram}.  
The gear reduction ratio of the WPG is
\begin{equation}\label{eq:wpg_ratio}
    G_{\text{WPG}} = \frac{1 + I_1}{1 - I_2},
\end{equation}
where
\begin{equation}\label{eq:wpg_I1}
    I_1 = \frac{N_{r_1}}{N_{s_1}}, \qquad
    I_2 = \frac{N_{r_1} N_{p_2}}{N_{p_1} N_{r_2}}.
\end{equation}

Here, $N_{s_1}$ and $N_{r_1}$ are the numbers of teeth on the sun gear $S$ and ring gear $R_1$, respectively, $N_{p_1}$ and $N_{p_2}$ are the teeth numbers of planets $P_1$ and $P_2$, and $N_{r_2}$ is the number of teeth on the output ring gear $R_2$.  

The efficiency of the WPG is
\begin{equation}\label{eq:wpg_eff}
    \eta_{\text{WPG}} = 
    \frac{\eta_{p_2r_2}\left(\eta_{p_1r_1} + \eta_{s_1p_1} I_1\right)(1 - I_2)}
    {(1 + I_1)\left(\eta_{p_1r_1}\eta_{p_2r_2} - I_2\right)},
\end{equation}
where $\eta_{\text{WPG}}$ is the overall efficiency, $\eta_{s_1p_1}$ the sun--planet ($S$--$P_1$) mesh efficiency, $\eta_{p_1r_1}$ the planet--ring ($P_1$--$R_1$) mesh efficiency, and $\eta_{p_2r_2}$ the planet--ring ($P_2$--$R_2$) mesh efficiency.

\subsection{Actuator Mass Model}\label{sec:mass_model}

To obtain actuator masses close to real values, we developed a detailed model accounting for every component. Each part is decomposed into simple geometries (cylinders, shells, cones). For example, a planetary gear in the CPG is modeled as two cylinders, with diameters and heights corresponding to the pitch circle diameters and face widths of Planet~1 and Planet~2. Using this approach, all components are approximated by suitable geometry combinations, as detailed in our repository. Predicted masses are compared with CAD-based estimates in Fig.~\ref{fig:mass_comparison}. Bearings, nuts, and bolts are selected from standard datasheets for accessibility and ease of assembly.

\section{Optimization of the Gearbox}\label{sec:opt}

This section formulates the gearbox optimization problem. 

\subsection{Optimization Variables}\label{subsec:Opt_var}
A unified notation is used for all gearbox types (SSPG, CPG, WPG, DSPG). 
The design variables are common across configurations and collected in vector 
$X \in \mathcal{X} \subseteq \mathbb{R}^{10}$:
\begin{equation}
    X := [N^s_1, N^p_1, N^r_1, N^s_2, N^p_2, N^r_2, m_1, m_2, n^p_1, n^p_2],
\end{equation}
where $N^s_i, N^p_i, N^r_i$ are the sun, planet, and ring gear teeth counts, 
$m_i$ is the module, and $n^p_i$ the number of planets in stage $i$ ($i \in \{1,2\}$). 
All variables are integers except $m_i$, which is chosen from a discrete set.  
Not all gearboxes use both stages. In SSPG, stage~2 is absent 
($N^s_2, N^p_2, N^r_2, m_2, n^p_2=0$). In CPG, $N^r_1=0$ and $N^s_2=0$. 
In WPG, $N^s_2=0$. In both CPG and WPG, planets are rigidly attached, so $n^p_1=n^p_2$. 
DSPG has no such restrictions.

\subsection{Constraints}
To ensure feasibility and manufacturability, the optimization problem includes several classes of constraints, described below.

\begin{enumerate}[label=\Roman*.]

\item \textbf{Gear Ratio Constraint:} The gear ratio of all the types of gearboxes are discussed in Section \ref{sec:gb_des}.
The following constraint ensures it remains within the specified range:
\begin{equation}\label{eq:gear_ratio_constr}
    GR_{min} \leq GR\leq GR_{max}
\end{equation}

\item \textbf{Geometric Constraints:}  
These constraints enforce dimensional compatibility between the sun, planet, and ring gears within each stage. The conditions~\cite{sspgTeethMatchingCond} for different gearbox types are:  

\begin{equation}
\begin{aligned}
    \text{SSPG/DSPG:} \quad & N^r_i = N^s_i + 2N^p_i, \\[4pt]
    \text{CPG:} \quad & m_2 N^r_2 = m_1(N^s_1 + N^p_1) + m_2 N^p_2, \\[4pt]
    \text{WPG:} \quad &
        \begin{aligned}[t]
            N^r_1 &= N^s_1 + 2N^p_1, \\
            m_2 N^r_2 &= m_1(N^s_1 + N^p_1) + m_2 N^p_2, \\
            m_2 N^r_2 &< m_1 N^r_1.
        \end{aligned}
\end{aligned}
\end{equation}

\begin{figure}[htbp]
    \centering
    \includegraphics[width=0.8\linewidth]{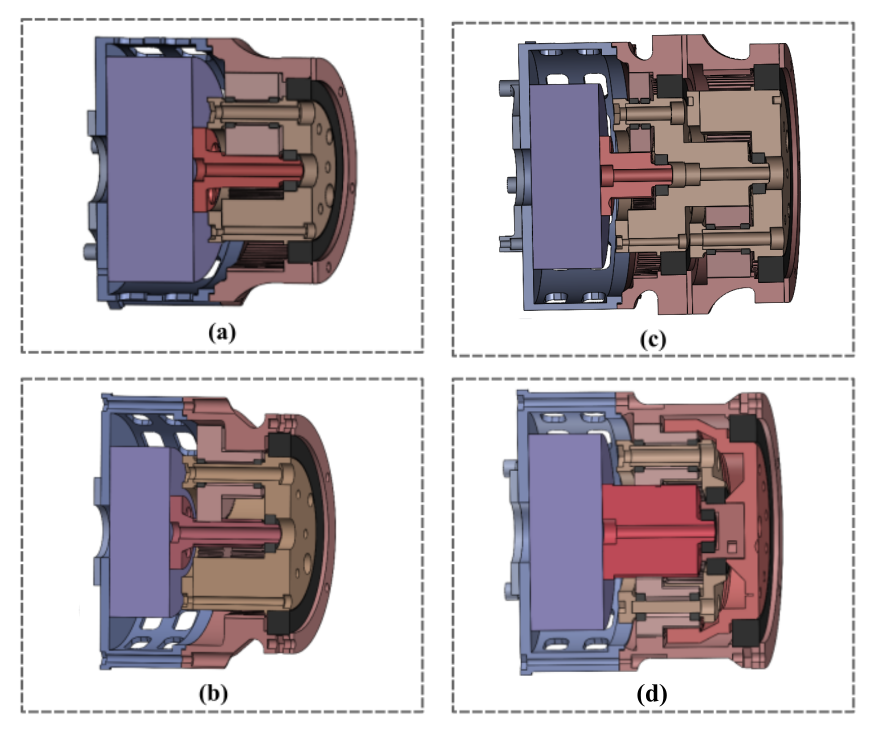}
    \caption{3D CAD models of the actuators (Red: \textbf{gear assembly}, Blue: \textbf{motor assembly}). 
(a) \textbf{SSPG:} 7.2:1, Motor: T-motor MN8014;
(b) \textbf{CPG:} 14:1, Motor: MAD-M6C12; 
(c) \textbf{DSPG:} 22.05:1, Motor: MAD-M6C12;
(d) \textbf{WPG:} 16.36:1, Motor: MAD-M6C12. Refer Table~\ref{tab:actuator_designs} for specifications.}
    \label{fig:3dp_cad}
\end{figure}

The additional inequality in the WPG ensures that the stage-1 ring gear is larger than the stage-2 ring gear. This configuration enables higher gear reduction for the same available space, thereby achieving a more torque-dense design (see Section~\ref{subsec:wpg_gb}).

\item \textbf{Meshing Constraints:}  
These constraints enforce proper tooth engagement between gears in each configuration~\cite{sspgTeethMatchingCond}. They are expressed using the modulo operator, where $a \ \% \ b = 0$ indicates that $a$ is divisible by $b$:  

\begin{equation}
\begin{aligned}
    \text{SSPG/DSPG:} \quad & (N^s_i + N^r_i) \ \% \ n^p_i = 0, \\[4pt]
    \text{CPG:} \quad & N^s_1 \ \% \ n^p_1 = 0, \quad N^r_2 \ \% \ n^p_1 = 0, \\[4pt]
    \text{WPG:} \quad &
        \begin{aligned}[t]
            (N^s_1 + N^r_1) \ \% \ n^p_1 &= 0, \\
            N^s_1 \ \% \ n^p_1 &= 0, \\
            N^r_2 \ \% \ n^p_1 &= 0.
        \end{aligned}
\end{aligned}
\end{equation}

For CPG and WPG, the planets are rigidly connected across stages, hence $n^p_1 = n^p_2$.

\item \textbf{No Interference Constraints:}  
These constraints prevent collisions between adjacent planet gears and the carrier extrusion. The conditions are formulated using pitch radii~\cite{sspgTeethMatchingCond}:  

\begin{equation}
\begin{aligned}
    \text{SSPG/DSPG:} \ & 
    2(R^s_{i} + R^p_{i}) \sin\!\left(\tfrac{\pi}{2n^p_{i}}\right) 
    - R^p_{i} - R_{ce} \geq \delta_{ce}, \\[6pt]
    \text{CPG/WPG:} \ & 
    2(R^s_{1} + R^p_{1}) \sin\!\left(\tfrac{\pi}{2n^p_{1}}\right) 
    - R^p_{1} - R_{ce} \geq \delta_{ce}.
\end{aligned}
\end{equation}

Here, $R^s_i$ and $R^p_i$ denote the pitch radii of the sun and planet gears in stage~$i$, $R_{ce}=4\text{ mm}$ is the carrier extrusion radius, and $\delta_{ce}=1\text{ mm}$ is the minimum clearance to avoid interference. The carrier extrusion connects the front (primary) and rear (secondary) carriers and lies between adjacent planet gears.

\item \textbf{Maximum Outer Gearbox Diameter Constraints:}  
These constraints limit the maximum outer diameter of the gearbox. For SSPG, WPG and DSPG, the ring gear diameter must not exceed $D^{\max}_{GB}$, while in CPG the outermost point of the larger planet gear must remain within $D^{\max}_{GB}$. 

\begin{equation}\label{eq:max_gb_od}
\begin{aligned}
    \text{SSPG/DSPG/WPG:} \quad & m_i N^r_i + \delta_{rw} \leq D^{\max}_{GB}, \\[6pt]
    \text{CPG:} \quad & m_1 (N^s_1 + 2N^p_1) \leq D^{\max}_{GB}.
\end{aligned}
\end{equation}

Here, $\delta_{rw}$ is the radial width of the ring gear (taken as $5$ mm in this work). The maximum gearbox diameter is defined as $D^{\max}_{GB} = K_{mgd} D_{motor}$, where $K_{mgd}$ is a user-defined hyperparameter specifying the allowable gearbox-to-motor diameter ratio.

\item \textbf{Additional Constraints:}  
These constraints restrict the feasible ranges of the design variables:
\begin{equation}\label{eq:additional_constr}
\begin{split}
    m_{\min} \leq m_i \leq m_{\max}, \quad N^s_i, N^p_i \geq N_{\min}, \\
    n^{\min}_p \leq n^p_i \leq n^{\max}_p.
\end{split}
\end{equation}

Some variables are fixed to zero depending on the gearbox type (see Section~\ref{subsec:Opt_var}) and will not follow these constraints; their separate formulations are excluded for brevity. In this work, we use $m_{\min} = 0.5$ mm, $m_{\max} = 1.2$ mm, and $N_{\min} = 18$ to avoid undercutting. The number of planet gears is constrained to $n_p \in [2,7]$ following standard practice.

\end{enumerate}

\subsection{Optimization Formulation}
The objective is to minimize actuator mass, maximize gearbox efficiency, and reduce axial width while satisfying a target gear ratio or range. Actuator mass $M_{act}$ is computed as in Section~\ref{sec:mass_model}, and efficiency $\eta$ as in Section~\ref{sec:gb_des}. Gear widths are derived from the modified Lewis strength equation~\cite{Maitra2010GearDesign}, while overall actuator width $W_p$ is primarily set by gearbox width, since motor and bearing lengths remain fixed. $W_p$ is therefore included in the cost function to penalize excessive axial dimensions.
For a specific ratio $GR_{req}$, a penalty weighted by $K_g$ accounts for deviation between $GR$ and $GR_{req}$. Equivalently, bounds can be set as $GR_{\min}=GR_{req}-1$ and $GR_{\max}=GR_{req}+1$. For ratio ranges, $K_g=0$ with limits $[GR_{\min},GR_{\max}]$. The cost function is
\begin{equation}\label{eq:cost}
    C_{\text{act}} := K_m M_{act} - K_e \eta + K_w W_p + K_g |GR_{req} - GR|
\end{equation}
where $K_m$, $K_e$, $K_w$, and $K_g$ weight mass, efficiency, width, and ratio tracking.  
The optimization problem is
\begin{equation}
\begin{aligned}
    \min_{\mathcal{X}} \quad & C_{\text{act}} \\
    \text{s.t.}\quad & \text{Constraints I--VI}.
\end{aligned}
\end{equation}

The solution $C_{act}^*$ yields the minimum cost for a specified gear ratio or ratio range. Since all variables are discrete, the problem is solved using brute-force search by evaluating all feasible combinations in the constraint set. This guarantees the global optimum but is computationally more expensive than other methods. In practice, runtimes remain modest: SSPG solves in a fraction of a second, CPG and WPG in under five seconds, and DSPG within a few minutes for a given gear ratio.

\section{Automation of Actuator design}\label{sec:auto}

\subsubsection{Actuator Design}
Each actuator is modular, comprising a \textit{motor subassembly} (motor, casing, driver) and a \textit{gearbox subassembly} (gears and carriers, varying by type), as shown in Fig.~\ref{fig:3dp_cad}. To improve durability of 3D-printed parts, the sun gear shaft is reinforced with a central bolt, and the motor casing incorporates through-bolts that act as a structural skeleton, enhancing housing stiffness.

\subsubsection{Motor and Motor Driver Selection}
The COMPAct framework accepts different motors and drivers as inputs to generate CAD designs.  The optimization and automation pipeline uses the motor’s theoretical torque capacity to determine gear width, while motor and driver dimensions are treated as fixed design dimensions (see Section~\ref{subsec:auto}). In our study, we use T-Motor (U8, U10, U12, MN8014), MAD M6C12, and Vector 8020 motors with ODrive Pro and Moteus drivers. Additional motors and drivers can be integrated by providing their dimensions and torque–power data.

\subsubsection{Automation of Actuator Design}\label{subsec:auto}
The actuator design process was automated using \textit{parametric CAD modeling}, linking optimization variables directly to geometric dimensions. All design parameters were defined as CAD variables, updated by the optimization program, and grouped as: (i) \textbf{Optimization variables} (e.g., gear teeth, module, planet count); (ii) \textbf{Fixed dimensions} (motor casing, clearances, hole sizes); and (iii) \textbf{Dependent dimensions} (functions of the first two).  
Optimization yields the optimal teeth numbers, module, and planet count, which update the CAD model. Together with fixed dimensions, they define the dependent dimensions, ensuring automatic updates. The parametric model was designed to avoid interference and assembly conflicts across gear ratios and configurations.

\subsubsection{Integration with Optimization Framework}
The communication between the optimization algorithm and the CAD model is managed through a \textit{text-based variable file}. All design variables, including optimization, fixed, and dependent, are stored in this file, which is linked to the CAD environment. Whenever the optimization program updates the file, the CAD model is automatically refreshed to reflect the new design. For the four gearbox architectures investigated, namely SSPG, CPG, WPG, and DSPG, the total number of design variables defined in the framework is approximately \textbf{150 for each type}. 

\subsubsection{Customization and Hyperparameters}
The automation framework also allows user-level customization through both fixed design parameters and optimization hyperparameters. Examples of fixed design variables include motor outer casing thickness, clearances between components, mounting plate thickness, and hole diameters. In addition, hyperparameters such as the maximum allowable gearbox diameter, motor selection, motor driver configuration, and variable bounds for optimization can be modified to suit different design requirements. 

\begin{figure}[htbp]
    \centering
    \includegraphics[width=0.85\linewidth]{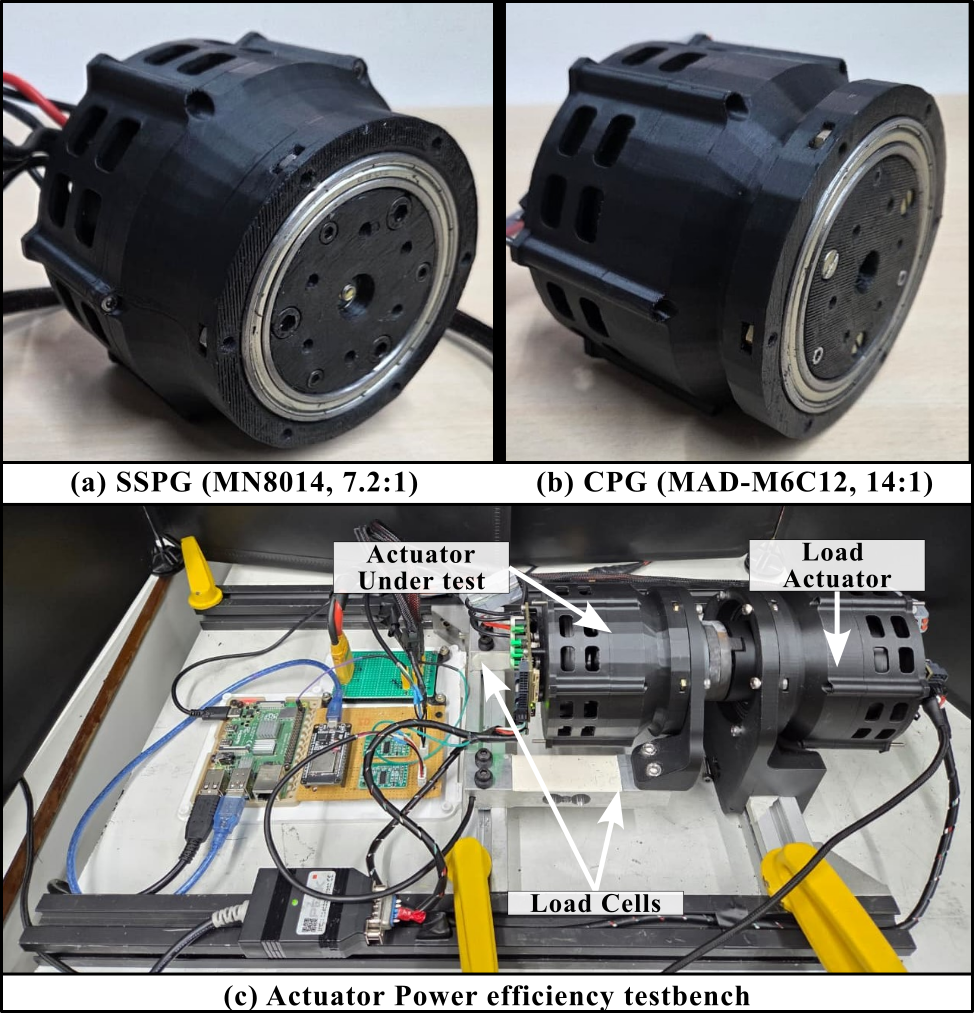}
    \caption{(a) \textbf{Assembled SSPG actuator}, (b) \textbf{Assembled CPG actuator}, (c) \textbf{Power efficiency testbench}: two load cells measure the torque of the actuator under test, while a second actuator serves as the load.}
    \label{fig:Actuator_and_testbench}
\end{figure}

\begin{figure}[htbp]
    \centering
    \includegraphics[width=0.8\linewidth]{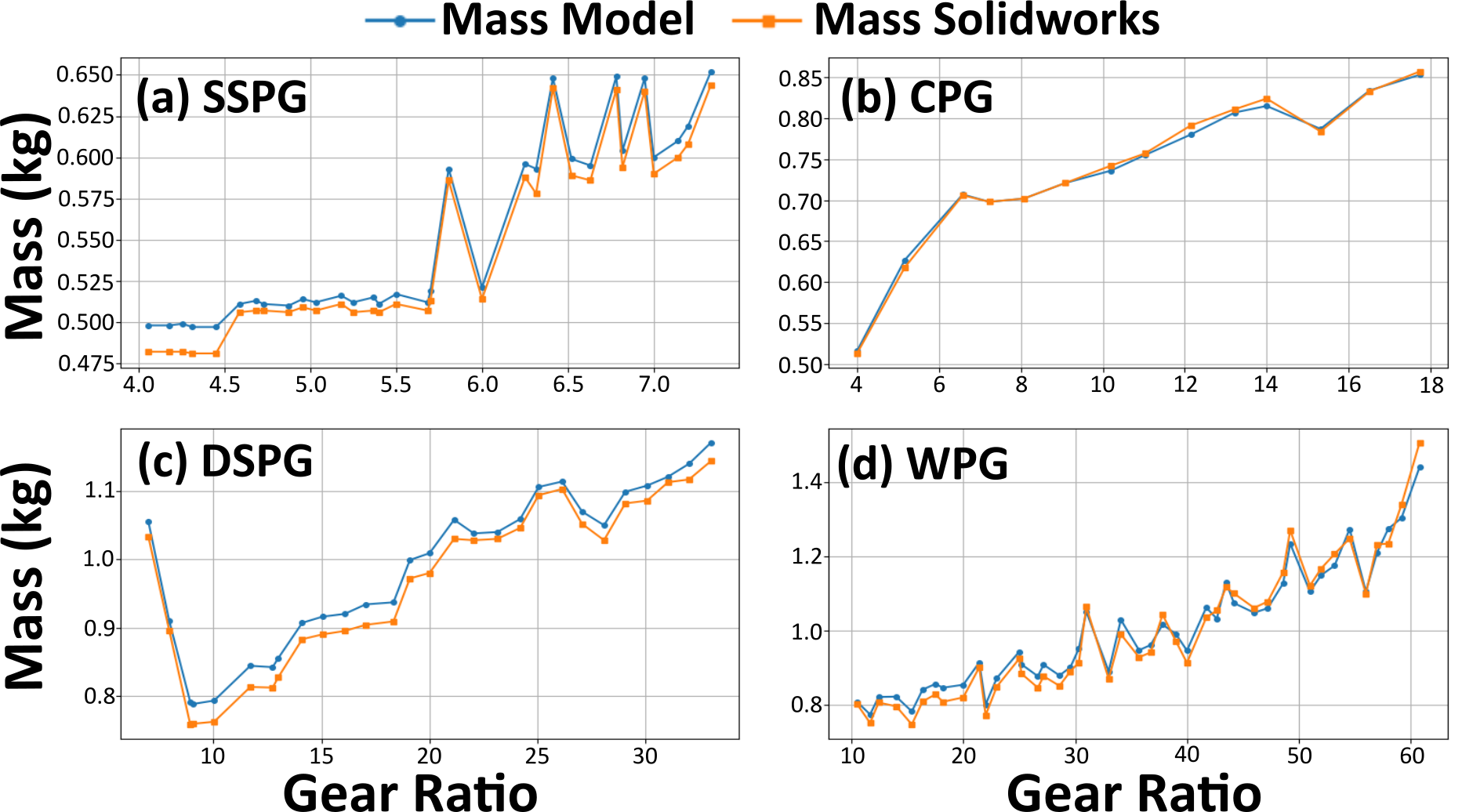}
    \caption{Comparison of mass model predictions with 3D CAD estimates. The model closely matches the final CAD masses, with RMS errors of 0.01 kg (SSPG), 0.005 kg (CPG), 0.024 kg (DSPG), and 0.028 kg (WPG).}
    \label{fig:mass_comparison}
\end{figure}

\begin{figure*}[htbp]
    \centering
    \includegraphics[width=0.8\linewidth]{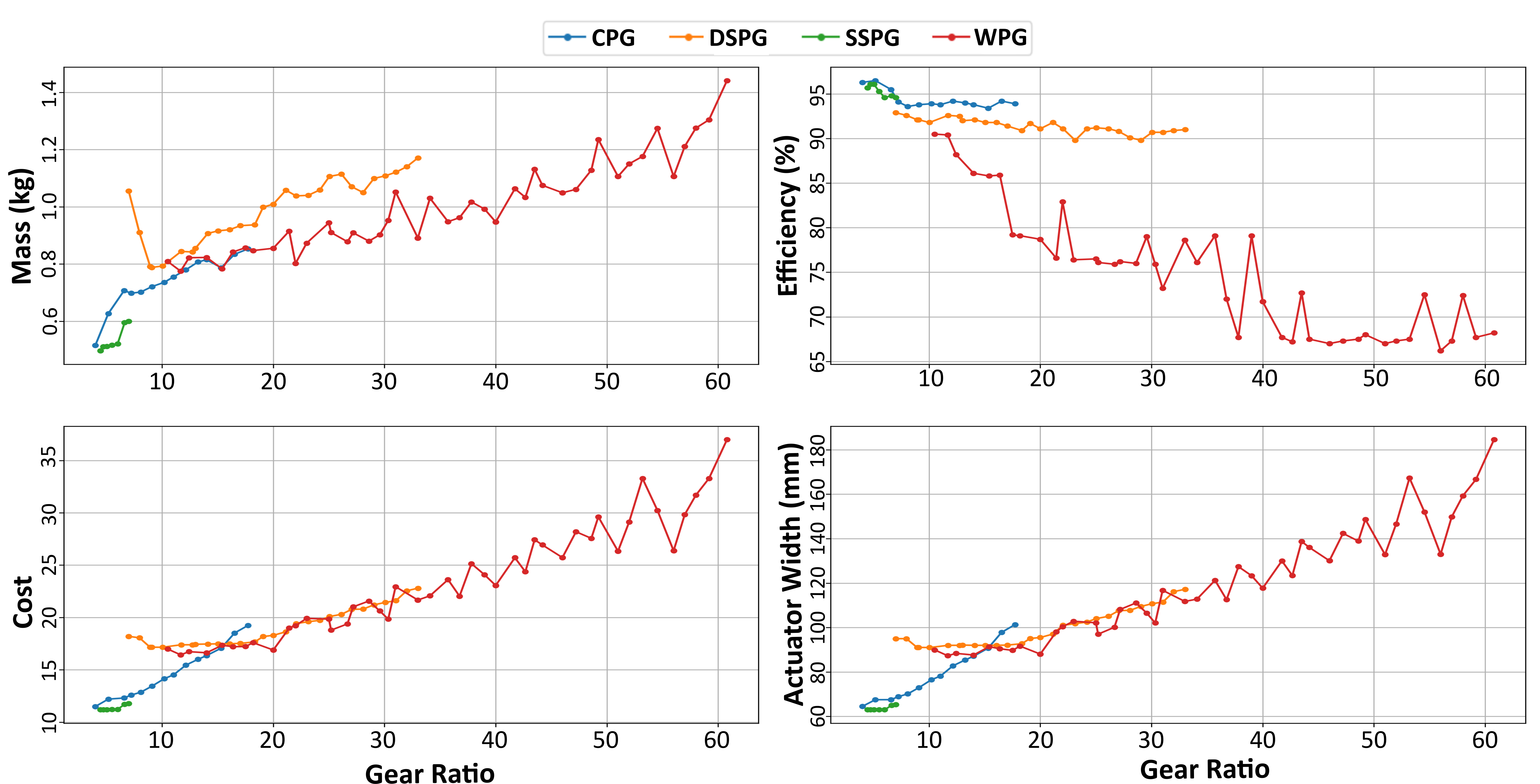}
    \caption{Optimization results for gearbox ratios between 4:1–60:1 using the MAD-M6C12 motor. Mass, efficiency, width, and cost are plotted against gear ratio to identify the optimal gearbox type. Here, $K_m=1$, $K_e=-2$, $K_w=0.2$, \& $K_g=0$ (ratio-range optimization).}
    \label{fig:GB_comparison}
\end{figure*}

\begin{figure}[htbp]
    \centering
    \includegraphics[width=0.85\linewidth]{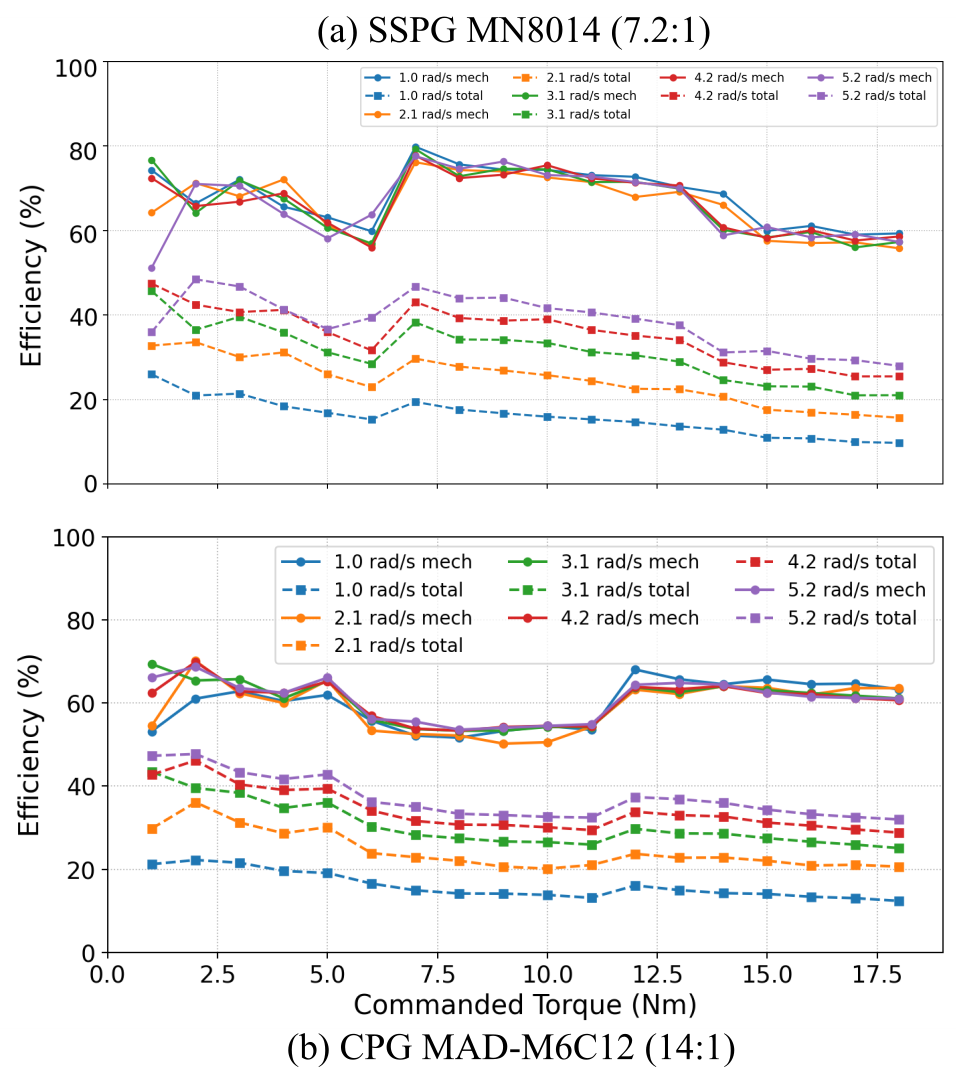}
    \caption{Efficiency of SSPG (top) and CPG (bottom). Mechanical efficiency (solid lines) is measured power over commanded torque–velocity product, while total efficiency (dotted lines) is measured power over electrical input power.}
    \label{fig:power_eff}
\end{figure}

\section{Results}\label{sec:results}

\subsection{Simulation Results}

This study evaluates the capability of the proposed framework to identify optimal actuator parameters and gearbox types for a given motor. The MAD M6C12 motor was used for the analysis. Optimal gearbox designs were generated for gear ratios ranging from 4:1 to 60:1 in unit increments (e.g., 4:1--5:1, 5:1--6:1, \dots, 59:1--60:1) across four gearbox types: SSPG, CPG, DSPG, and WPG. The actuator mass, efficiency, width, and cost were computed for each case, and the results are summarized in Figs.~\ref{fig:mass_comparison} and \ref{fig:GB_comparison}.

\subsubsection{Validation of the Mass Model}  
The accuracy of the actuator mass model used in the optimization framework was assessed by comparing its predictions against masses obtained from automated parametric CAD models. The root-mean-square (RMS) error between the two estimates was computed for each gearbox type.  
As shown in Fig.~\ref{fig:mass_comparison}, the difference between the model and CAD-based calculations is minimal. The RMS error for SSPG, CPG, DSPG and WPG is $0.01$ kg, $0.005$ kg, $0.024$ kg, and $0.028$ kg, respectively. These results confirm that the proposed mass model closely approximates the real actuator mass, thereby strengthening the reliability of the optimization framework.  

\subsubsection{Optimal Gearbox Designs}  
The framework was further used to compare gearbox types and identify the most suitable designs across the specified ratio range. The following observations can be made from Fig.~\ref{fig:GB_comparison}:

\begin{enumerate}
    \item The feasible design space varies significantly. SSPG is limited to ratios up to 7.2:1, CPG to 18:1, DSPG to 32:1, while only WPG spans the full 4:1--60:1 range.
    \item In the 4:1--7.2:1 range, SSPG offers the lowest cost, while also being the lightest, most efficient, and most compact actuator.
    \item Between 7.2:1 and 15:1, CPG emerges as the optimal option, with a cost advantage and efficiency nearly comparable to SSPG.
    \item In the 15:1--32:1 range, WPG and DSPG are closely competing, with DSPG showing higher efficiency. Beyond 32:1, only WPG remains feasible.
\end{enumerate}

These results highlight the utility of the COMPAct framework in validating design models and supporting informed actuator selection across a wide range of applications. For example, when selecting a gearbox with the MAD-M6C12 motor in the 14--15:1 ratio range, the framework identifies the CPG as the optimal solution (minimum cost), with the corresponding parameters listed in Table~\ref{table:Opt_param}. The framework also automates CAD generation of the selected design, as illustrated in Fig.~\ref{fig:3dp_cad}.  
While the framework produces detailed optimal parameters for all gearboxes, only representative results for selected cases are reported in Table~\ref{table:Opt_param} for brevity.

\subsection{Hardware Results}
Using the automation framework, four optimal actuators are designed (Fig.~\ref{fig:3dp_cad}). The SSPG and CPG actuators are 3D-printed and experimentally evaluated. Both actuators are fabricated using identical printing settings with $100\%$ infill density on a Bambu Lab X1C printer. The \textbf{SSPG actuator} uses an MN8014 motor ($87.8~\text{mm}$ OD), whereas the \textbf{CPG actuator} employs a MAD M6C12 motor ($72~\text{mm}$ OD). Accordingly, the hyperparameter $K_{mgd}$ in Eq.~(\ref{eq:max_gb_od}) is set to $1$ for SSPG and $1.25$ for CPG.

\subsubsection{Power Efficiency Test}  
The power efficiency of the actuators was evaluated on a custom-built dynamometer, similar to the setup in Berkeley Humanoid Lite~\cite{berkeleyhumanoidlite} (Fig.~\ref{fig:Actuator_and_testbench}). The test actuator was run in torque-control mode, while the load actuator was velocity-controlled to maintain constant speed.  
Electrical input power was computed as supply voltage times current, and mechanical output power as output angular velocity times measured torque. Angular velocity was obtained from the motor encoder (scaled by gear reduction), and torque from load cells integrated into the mount.  
Mechanical efficiency, representing gearbox transmission efficiency, was defined as mechanical output power over commanded torque–velocity product. Overall actuator efficiency was defined as mechanical output power over electrical input power, thus accounting for motor Joule losses, driver efficiency, and bearing losses.  
Tests were performed at five angular velocities and torques up to 18~Nm. For each pair, the actuator was ramped to the target and held for 10s, with analysis based on steady-state data.
\begin{figure}[htbp]
    \centering
    \includegraphics[width=0.85\linewidth]{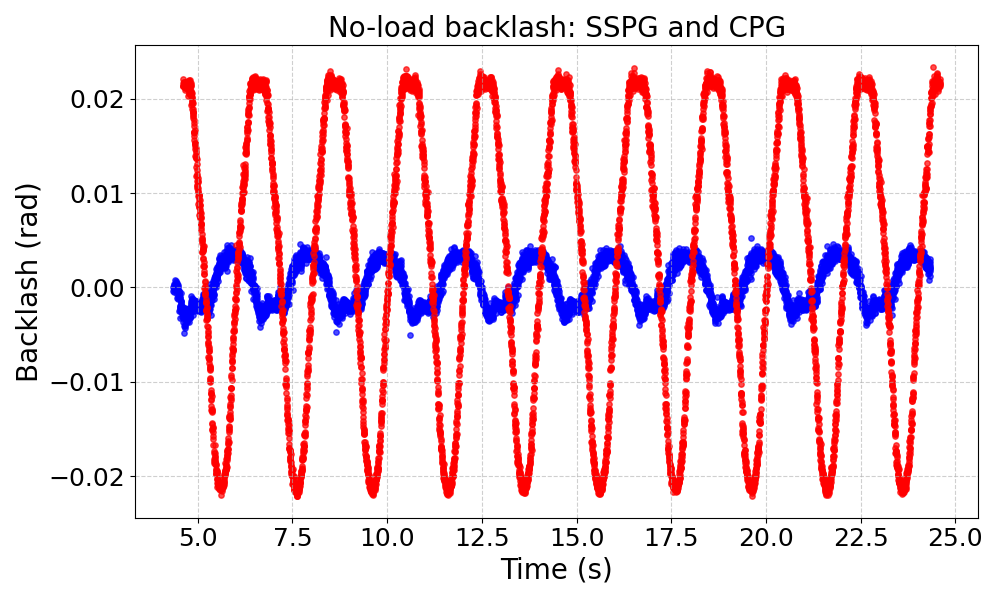}
    \caption{No-load backlash test results. The compound planetary gearbox (CPG, red) exhibits a backlash of $0.046~\text{rad}$ ($2.6^{\circ}$), whereas the single-stage planetary gearbox (SSPG, blue) shows a lower backlash of $0.010~\text{rad}$ ($0.59^{\circ}$)}
    \label{fig:no_load_backlash}
\end{figure}
\begin{table}[t]
\centering
\caption{Optimized Actuator Designs from COMPAct}
\label{tab:actuator_designs}
\begin{tabular}{lcl}
\hline
\textbf{Actuator (Motor)} & \textbf{Gear Ratio } & 
\textbf{Optimal Parameters} \\
 & {(Range)} & $[N^s_i, N^p_i, N^r_i, m_i, n^p_i]$ $i \in \{1,2\}$\\
\hline
SSPG (MN8014) & $7.2{:}1$ & 
\textbf{Stage-1:} $[25, 65, 155, 0.5, 3]$ \\
               & ($7.2$--$7.3{:}1$) & 
\textbf{Stage-2:} $[0, 0, 0, 0, 0]$ \\
\hline
CPG (M6C12)   & $14{:}1$ & 
\textbf{Stage-1:} $[18, 66, 0, 0.6, 3]$ \\
               &    ($14$--$15{:}1$)  & 
\textbf{Stage-2:} $[0, 33, 117, 0.6, 3]$ \\
\hline
DSPG (M6C12)  & $22.05{:}1$ & 
\textbf{Stage-1:} $[35, 52, 139, 0.5, 3]$ \\
               &  ($22$--$23{:}1$) & 
\textbf{Stage-2:} $[23, 28, 79, 1.0, 3]$ \\
\hline
WPG (M6C12)   & $16.36{:}1$ & 
\textbf{Stage-1:} $[66, 45, 156, 0.5, 6]$ \\
               &  ($16$--$17{:}1$)  & 
\textbf{Stage-2:} $[0, 33, 144, 0.5, 6]$ \\
\hline
\end{tabular}\label{table:Opt_param}
\vspace{1mm}

\footnotesize{
\textit{Notes:}  
1) Zero values indicate non-existent gears/parameters in that stage.  
2) In CPG and WPG, the two planet sets ($p_1, p_2$) are rigidly connected, hence the number of planets is equal across stages. 

3) $K_m=1$, $K_e=-2$, $K_w=0.2$, \& $K_g=0$ (ratio-range optimization)}
\end{table} 
The efficiency results are presented in Fig.\ref{fig:power_eff}.
The optimization framework predicted efficiencies of $96\%$ for SSPG and $93.8\%$ for CPG. Experimentally, the mechanical efficiency was lower: $60$--$80\%$ for SSPG and about $60\%$ for CPG. This discrepancy is attributed to deviations from the ideal involute tooth profile caused by 3D-printing resolution limits. Also, the lack of industrial-grade precision in 3D-printed parts introduces additional friction in the gear system. Nonetheless, SSPG consistently outperformed CPG, consistent with theoretical predictions.  
Total efficiency increased with angular velocity, while mechanical efficiency remained velocity-independent. At higher torques, total efficiency decreased, likely due to greater Joule losses.

\subsubsection{No-Load Backlash Test}
The no-load backlash of SSPG (7.2:1) and CPG (14:1) was measured using two encoders: the motor-side encoder (integrated in the driver) and a high-resolution output encoder. A low-frequency, small-amplitude sinusoidal velocity ensured smooth motion while neglecting gearbox compliance. The motor angle $\theta_m$ was scaled by the gear ratio $G$ to compare with the output angle $\theta_o$, and the relative motion was computed as $\Delta\theta(t) = \tfrac{\theta_m(t)}{G} - \theta_o(t)$. Backlash was estimated as $\beta = \max_t \Delta\theta(t) - \min_t \Delta\theta(t)$.  

The measured values were $\beta_{\text{SSPG}} = 0.010~\text{rad }(0.59^{\circ})$ and $\beta_{\text{CPG}} = 0.046~\text{rad }(2.6^{\circ})$, as shown in Fig.~\ref{fig:no_load_backlash}. The CPG exhibited higher backlash, likely due to the smaller effective planet diameters in its second stage, which reduce contact ratio and increase clearance.


\begin{figure}[htbp]
    \centering
    \includegraphics[width=0.85\linewidth]{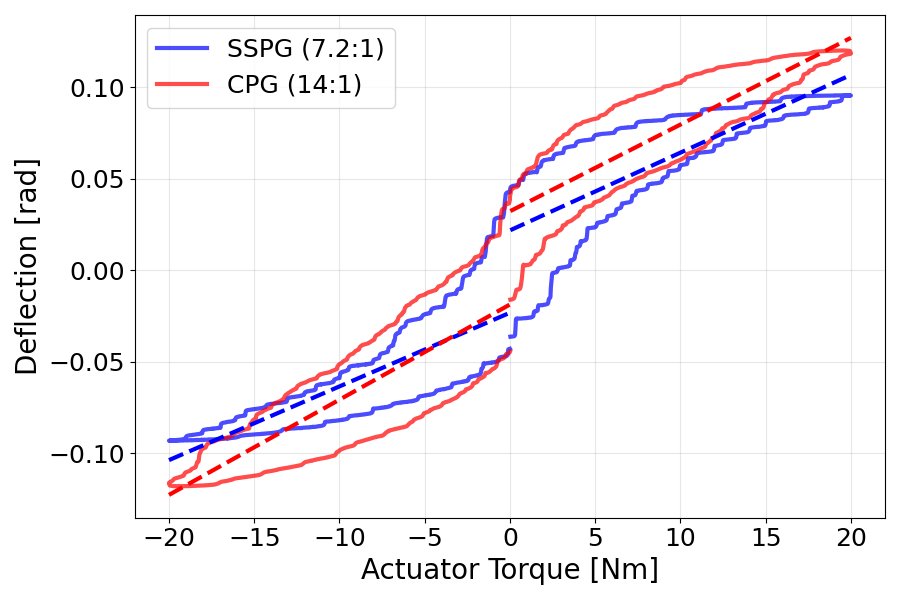}
    \caption{\textbf{Transmission stiffness:} The x-axis shows the commanded actuator torque at the output, and the y-axis indicates the internal deflection. The CPG (red) exhibits lower stiffness than the SSPG (blue).}
    \label{fig:stiffness_test}
\end{figure}

\begin{table}[!t]
\caption{Estimated stiffness values}
\label{tab:stiffness}
\centering
\begin{tabular}{l|ccc}
\hline
 & Forward & Backward & \textbf{Mean ($K$)} \\
\hline
SSPG (7.2:1) & 238.4 & 247.0 & \textbf{242.7 [Nm/rad]} \\
CPG (14:1)   & 211.0 & 192.2 & \textbf{201.6 [Nm/rad]} \\
\hline
\end{tabular}
\end{table}

\subsubsection{Transmission Stiffness}
Transmission stiffness was measured for SSPG (7.2:1) and CPG (14:1). The actuator output was rigidly fixed, and a motor torque command was applied: ramping from $0$ to $\pm20~\mathrm{Nm}$ and back. Motor encoder readings were divided by the gear ratio to obtain internal deflection. For each actuator, linear fits were applied separately to the positive and negative torque samples, $\theta = \tfrac{\tau}{K} + b$, where $\theta$ is deflection, $\tau$ the applied torque, $K$ the stiffness, and $b$ a bias term. The reported stiffness is the mean of the two estimates.  
Results (Fig.~\ref{fig:stiffness_test}) show hysteresis, consistent with prior reports on 3D-printed transmissions in Humanoid Lite~\cite{berkeleyhumanoidlite} and Roozing et. al.~\cite{Roozing_and_roozing}. Measured stiffness values (Table~\ref{tab:stiffness}) were $242.7~\mathrm{Nm/rad}$ for SSPG and $201.6~\mathrm{Nm/rad}$ for CPG. For comparison, Roozing~\cite{Roozing_and_roozing} reported $1468~\mathrm{Nm/rad}$ and Humanoid Lite~\cite{berkeleyhumanoidlite} $319~\mathrm{Nm/rad}$. Higher values in these works likely result from cycloidal transmissions (larger contact areas) and stronger materials (e.g., carbon-fiber–reinforced polyamide versus PLA in our prototypes). In our tests, CPG stiffness was lower than SSPG, attributed to the smaller planet diameter in its second stage, which reduces contact ratio and effective stiffness.

\section{Conclusion}\label{sec:conclusion}

This paper presented \textbf{COMPAct}, a framework for computational optimization and automated modular design of 3D-printed planetary actuators. By combining gearbox parameter optimization with CAD automation, the framework enables rapid design exploration and fabrication. Simulations identified feasible designs across four gearbox types: SSPG, CPG, DSPG, and WPG, while comparison with CAD models validated the mass model. Automatic CAD generation further enables seamless transition from optimization to fabrication. 
To demonstrate practicality, two actuators were fabricated and tested for power efficiency, transmission stiffness, and backlash, confirming the framework’s ability to produce lightweight and manufacturable designs. 

Future work will extend the framework to additional gearbox types, conduct endurance and peak torque tests, and improve efficiency via better torque sensing. Metal actuators and advanced optimization (e.g., MINLP solvers) will be incorporated, and rotor inertia will be added to the cost function.

\bibliographystyle{IEEEtran}
\bibliography{references}

\end{document}